\documentclass{article}

\PassOptionsToPackage{numbers, compress}{natbib}
 \usepackage[preprint]{neurips_2026}

\usepackage[utf8]{inputenc} % allow utf-8 input
\usepackage[T1]{fontenc}    % use 8-bit T1 fonts
\usepackage{hyperref}       % hyperlinks
\usepackage{url}            % simple URL typesetting
\usepackage{booktabs}       % professional-quality tables
\usepackage{amsfonts}       % blackboard math symbols
\usepackage{nicefrac}       % compact symbols for 1/2, etc.
\usepackage{microtype}      % microtypography
\usepackage{xcolor}         % colors
\usepackage{enumitem}
\usepackage{wrapfig}

\usepackage{graphicx}
\usepackage{subcaption}
\usepackage{tabularx}
\usepackage{makecell}
\usepackage{hyperref}
\usepackage{multirow}

\usepackage{amsmath}
\usepackage{amssymb}
\usepackage{mathtools}
\usepackage{amsthm}

\title{MemoryFusion: Cross-Temporal Memory Learning for Multimodal Video Fusion}

\author{%
  Meiqi Gong \\
  Electronic Information School \\
  Wuhan University \\
  Wuhan, China \\
  \texttt{meiqigong@whu.edu.cn} \\
  \And
  Hao Zhang$^{*}$ \\
  Electronic Information School \\
  Wuhan University \\
  Wuhan, China \\
  \texttt{zhpersonalbox@gmail.com} \\
  \And
  Jiayi Ma$^{*}$ \\
  School of Robotics \\
  Wuhan University \\
  Wuhan, China \\
  \texttt{jyma2010@gmail.com} \\
}

\begin{document}

\maketitle
\begingroup
\renewcommand\thefootnote{}\footnotetext{%
  \textsuperscript{*}\;Corresponding authors.
}
\addtocounter{footnote}{-1}
\endgroup

\begin{abstract}
Multimodal videos contain rich temporal cues that can facilitate faithful fusion. However, existing video fusion methods mainly rely on short-range temporal interactions, leaving valuable long-term information underexplored. Moreover, inconsistent fusion representations can cause flickering artifacts, undermining temporal coherence. To address these coupled challenges, we propose MemoryFusion, a two-stage framework integrating cross-temporal memory learning with progressive temporal refinement. Considering the complementary roles of adjacent and distant frames, the first stage combines a Short-Term Memory Bank for local temporal guidance with a Long-Term Memory Bank that preserves reliable historical information through a fusion-oriented update strategy. To mitigate temporal fluctuations, a Temporal Smoothing Module leverages soft motion cues for coarse background stabilization. Building upon the resulting fusion prior, The second stage further employs a lightweight 3D residual Temporal Refinement Module to correct remaining temporal inconsistencies while enhancing structural details. Extensive experiments demonstrate that MemoryFusion achieves superior fusion quality and temporal consistency over existing frame-based and video-based fusion methods. The source code is publicly available at \url{https://github.com/Meiqi-Gong/MemoryFusion}.
\end{abstract}

\section{Introduction}
Due to the inherent limitations of sensing mechanisms and hardware, single-modal images often fail to provide reliable scene understanding, especially in complex environments~\cite{gong2024cross,cao2023multi, zhang2025omnifuse}. To address this issue, multimodal image fusion has emerged as a crucial technique that integrates complementary information from different modalities to produce more comprehensive and reliable representations~\cite{xu2022cufd,li2024dstcfuse}. Benefiting from its enhanced representational capacity, multimodal fusion has been successfully applied in a wide range of fields, such as autonomous driving~\cite{xie2024dual,liu2023multi}, intelligent surveillance~\cite{zhang2024dispel}, and medical diagnosis~\cite{chen2025multi}, among others~\cite{liu2024coconet,luo2022thermal}.

\begin{figure}[t]
  \centering
  \hspace{-0.17in}
  \includegraphics[width=1\linewidth]{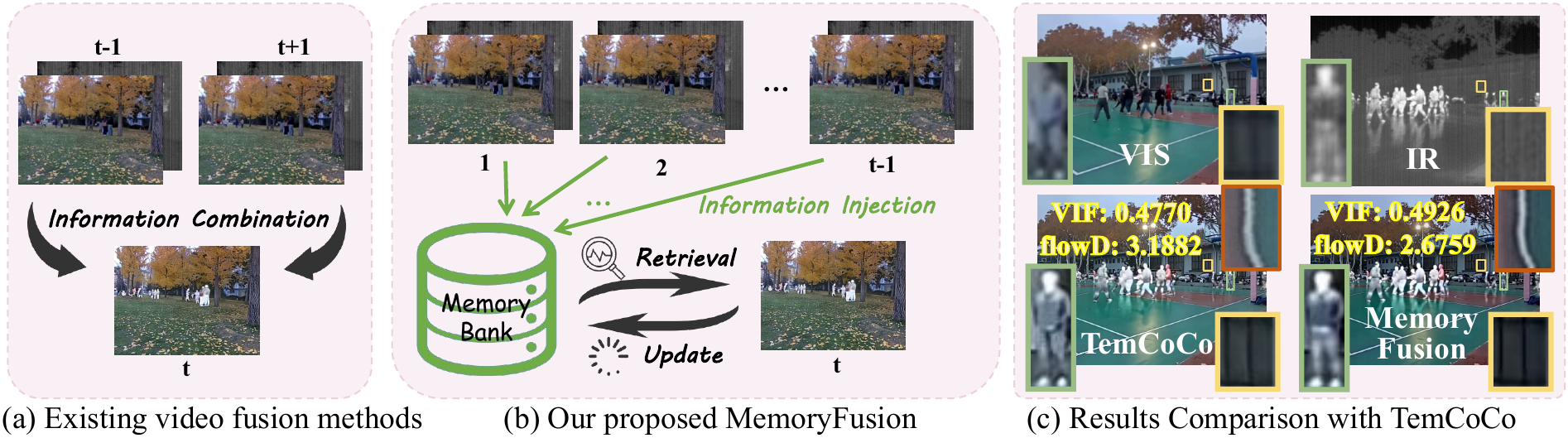}
  \vspace{-0.1in}
  \caption{Comparison of existing video fusion methods and our proposed MemoryFusion. (a) and (b) illustrate the differences in how temporal information is exploited. (c) shows the corresponding visual comparison results, where the top-right region displays the results of stacked 40 consecutive frames in the same row for temporal consistency evaluation. Higher VIF means better fusion quality and lower flowD means better temporal consistency.}
   \label{fig:intro}
\end{figure}

Early research on multi-modal fusion has primarily focused on frame-based design, where frame-based modality-specific features are typically combined to enhance detail and contrast. For instance, Ma \emph{et al.} introduced SwinFusion~\cite{ma2022swinfusion}, which employs a Swin Transformer backbone to jointly capture intra-modal and inter-modal spatial dependencies within a single frame, enabling more adaptive and content-aware frame-based fusion. After that, CDDFuse~\cite{zhao2023cddfuse} employs Restormer blocks to extract shallow cross-modal frame-wise features, followed by a dual-branch Transformer-CNN extractor that captures global and local representations within each frame independently. Further, Text-DiFuse~\cite{zhang2024text} introduces a text-modulated diffusion model into the frame-based fusion paradigm. It embeds feature-level integration directly into the single-frame generative diffusion process, enabling simultaneous degradation removal and cross-modal information fusion within each frame. Recently, FC-Former~\cite{wu2025fully} uses a fully-connected self-attention mechanism to capture local and non-local correlations within each single frame. It embeds a multi-source image representation as a non-local prior to enable effective single-frame feature fusion across modalities. Despite their remarkable success in image fusion, they are designed under single frame, whereas in practical scenarios, multimodal data are typically acquired in video form. This discrepancy prevents such methods from effectively modeling and leveraging inter-frame dependencies, thereby resulting in limited fusion performance and undesired temporal flickering when applied to practical video sequences.

Due to the growing recognition of the limitations of frame-based fusion, researchers have recently shifted their attention toward video fusion, which explicitly models and exploits temporal information across consecutive frames. By leveraging temporal correlations, video fusion enables a more comprehensive and stable fusion of multimodal data, resulting in significantly improved visual quality and reduced inter-frame flickering. For example, Gong \emph{et al.} proposed TemCoCo~\cite{gong2025temcoco}, which aligns features from adjacent frames through deformable convolution (DCN) to capture local short-term correlations, while incorporating a temporal modeling module and a dedicated temporal consistency loss to ensure temporal consistency. In addition, Tang \emph{et al.} introduced VideoFusion~\cite{tang2025videofusion}, which employs an attention mechanism to associate neighboring frames with the current one and maintains temporal coherence through explicit consistency constraints. Besides that, Zhao \emph{et al.} proposed UniVF~\cite{zhao2025unified}, leveraging neighboring-frame learning and optical flow-based feature warping to fuse adjacent frames. Also, they integrate a Transformer-based encoder-decoder with a temporal consistency loss to enhance temporal coherence. Overall, the development of video fusion methods is still at a preliminary stage. Existing video fusion methods primarily focus on short-range correlations between adjacent frames, as concluded in Figure \ref{fig:intro} (a), where information of adjacent frames is transferred through various strategies to assist current-frame fusion. This short-term dependency limits the full exploitation of accessible temporal context,  thus constraining fusion fidelity. Moreover, their temporal consistency modeling is confined to complex networks with merely simple temporal constraints and no incorporation of temporal or spatial prior, which often results in limited smoothness and the loss of fine spatial details. 

Beyond the video fusion literature, memory mechanisms have also been investigated in broader video processing tasks. For example, memory banks are commonly used in video object segmentation to establish long-term correspondence for mask propagation~\cite{zhou2024rmem,cheng2022xmem}, while motion-adaptive temporal filtering and 3D convolution have been explored to suppress temporal fluctuations and capture local spatio-temporal dependencies~\cite{peng2024video, ying2020deformable}. These studies demonstrate the potential of temporal memory and temporal filtering for video understanding. Inspired by these temporal modeling paradigms, we redesign them under the unique characteristics of multimodal video fusion.

To address the above limitation, we propose MemoryFusion, a two-stage multimodal video fusion method to explicitly tackle the challenges of underutilized temporal context and inadequate temporal consistency. On the one hand, we fully exploit cross-temporal learning through dual memory banks in the first stage. Specifically, the Short-Term Memory Bank (STMB) aggregates informative cues from recent frames to provide local temporal guidance, while the Long-Term Memory Bank (LTMB) adaptively captures and preserves key representations from long-term frames to supply global temporal guidance. As shown in Figure~\ref{fig:intro} (b), unlike conventional methods that rely solely on short-range temporal cues, our MemoryFusion also explicitly maintains long-term temporal information through dedicated memory banks, enabling the network to retrieve more comprehensive information. On the other hand, to alleviate temporal fluctuations introduced during cross-frame feature aggregation, a Temporal Smoothing Module (TSM) is incorporated at the end of the first stage to perform region-adaptive coarse temporal smoothing by selectively preserving background information from the preceding frame. Building upon the resulting fusion prior, the second stage incorporates a Temporal Refinement Module (TRM) to guarantee fine temporal coherence. Treating the first-stage output as the initial results, the TRM leverages a 3D convolutional network to explicitly model and enforce temporal constraints, achieving a holistic optimization. Through this progressive coarse-to-fine design, MemoryFusion effectively integrates historical information while maintaining temporal stability. As illustrated in Figure~\ref{fig:intro}(c), our MemoryFusion not only restores structural details that remain blurred in results from other video fusion methods such as TemCoCo, but also preserves temporal coherence across consecutive frames, effectively suppressing flickering artifacts and ensuring stable visual dynamics.

Our main contributions of this work can be summarized as:
\begin{itemize}[leftmargin=1.2em,labelsep=0.4em]
\item We propose MemoryFusion, a cross-temporal memory learning for multimodal video fusion that jointly enhances visual quality and temporal consistency.
\item Dual memory banks are designed with a redundancy-aware and fusion-quality-aware LTMB update strategy to preserve diverse and reliable historical information.
\item We develop a two-fold progressive temporal modeling approach, a TSM alleviating coarse background transitions, and a TRM enforcing fine consistency while refining details.
\item Extensive experiments on public datasets demonstrate that MemoryFusion achieves superior performance in terms of fusion quality and temporal consistency.
\end{itemize}

\begin{figure*}[t]
  \centering
   \includegraphics[width=0.98\linewidth]{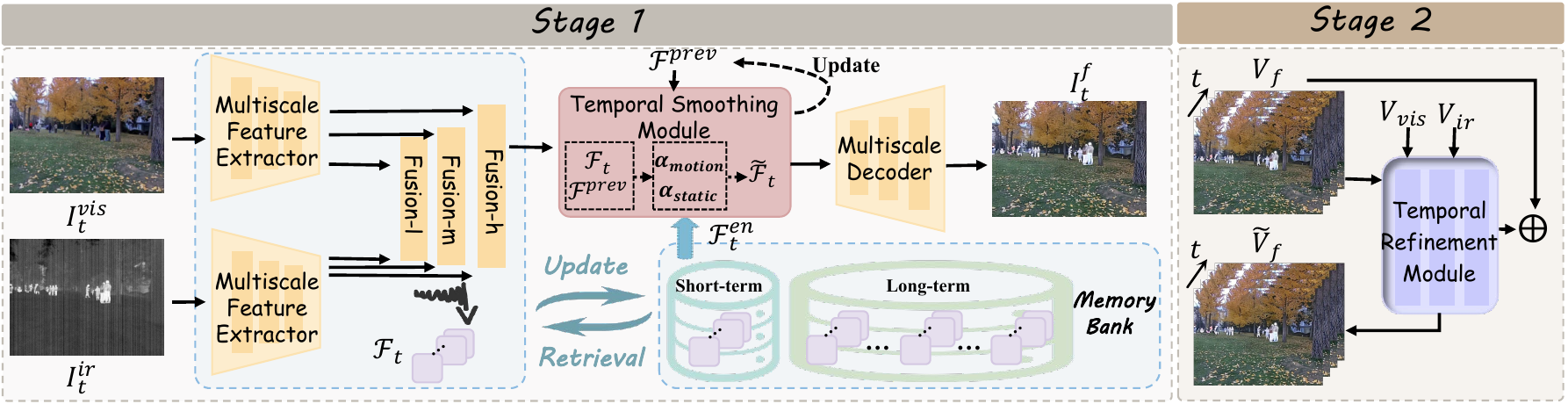}
   \caption{The overall framework of the proposed MemoryFusion.}
   \label{fig:overview}
\end{figure*}

\section{Method}
This section details the proposed two-stage MemoryFusion, as illustrated in Figure~\ref{fig:overview}. The first stage performs memory-guided fusion with coarse temporal consistency modeling, while the second stage refines temporal consistency based on the fusion prior from the first stage.

\subsection{Notation and Overview}
In the first stage, dual memory banks are designed to store information from multiple temporal scales, and the framework primarily focuses on exploiting historical information stored in the memory bank to facilitate enhanced video fusion. Taking the $t$-th frame as an example, the visible frame \(I_{t}^{\text{vis}}\in\mathbb{R}^{H\times W\times 3}\) and the infrared frame \(I_{t}^{\text{ir}}\in\mathbb{R}^{H\times W\times 1}\) are first fed into two Multiscale Feature Encoders $\mathcal{E}_{\text{vis}}, \mathcal{E}_{\text{ir}}$ to extract hierarchical representations by:
\begin{align}
\mathcal{F}_{t,l}^{\text{vis}},\mathcal{F}_{t,m}^{\text{vis}},\mathcal{F}_{t,h}^{\text{vis}} = \mathcal{E}_{\text{vis}}(I_{t}^{\text{vis}}),\quad
\mathcal{F}_{t,l}^{\text{ir}},\mathcal{F}_{t,m}^{\text{ir}},\mathcal{F}_{t,h}^{\text{ir}} = \mathcal{E}_{\text{ir}}(I_{t}^{\text{ir}}),
\end{align}
where $\mathcal{F}_{t,l}^{\text{vis}},\mathcal{F}_{t,l}^{\text{ir}}\in\mathbb{R}^{\frac{H}{4}\times \frac{W}{4}\times C_1}$, $\mathcal{F}_{t,m}^{\text{vis}},\mathcal{F}_{t,m}^{\text{ir}}\in\mathbb{R}^{\frac{H}{2}\times \frac{W}{2}\times C_2}$ and $\mathcal{F}_{t,h}^{\text{vis}},\mathcal{F}_{t,h}^{\text{ir}}\in\mathbb{R}^{H\times W\times C_3}$. The corresponding multi-modal features are then fused through multiple feature fusion layers to obtain preliminary fused features $\mathcal{F}_{t,l}^f, \mathcal{F}_{t,m}^f$ and $\mathcal{F}_{t,h}^f$. These fused features interact with the memory banks through a retrieval process, where the network selectively extracts the relevant information from the memory banks to support the current fusion. The memory banks are dynamically updated following distinct strategies: the Short-term Memory Bank (STMB) is refreshed at every frame, while the Long-term Memory Bank (LTMB) is updated only when its capacity limit is reached, following a specific update rule that will be detailed in Section~\ref{Sec:Memory}. Subsequently, the features are passed to the Temporal Smoothing Module (TSM), which integrates temporal information by incorporating the previous features $\mathcal{F}^{\text{prev}}$ at corresponding network layers to get the coarse temporal-smoothed feature $\mathcal{F}^{sm}_t$. Finally, the memory-enhanced features $\mathcal{F}^{\text{en}}_t$ and the smoothed fusion features $\mathcal{F}^{\text{sm}}_t$ are fed into a Multiscale Decoder $\mathcal{D}$ to reconstruct the fused frame $I^f_t$:
\begin{equation}
 I^f_t = \mathcal{D}(\mathcal{F}^{\text{en}}_t,\mathcal{F}^{\text{sm}}_t).
\end{equation}
Except for the first frame ($\mathcal{F}^{prev}$ is absent), all subsequent frames follow the same processing pipeline. After the first stage, the entire fused video sequence $V_f$, together with the original input video sequences $V_{vis}$ and $V_{ir}$, is fed into the second-stage Temporal Refinement Module (TRM) to further enhance global temporal consistency across the video.

\begin{figure}[t]
  \centering
   \includegraphics[width=1\linewidth]{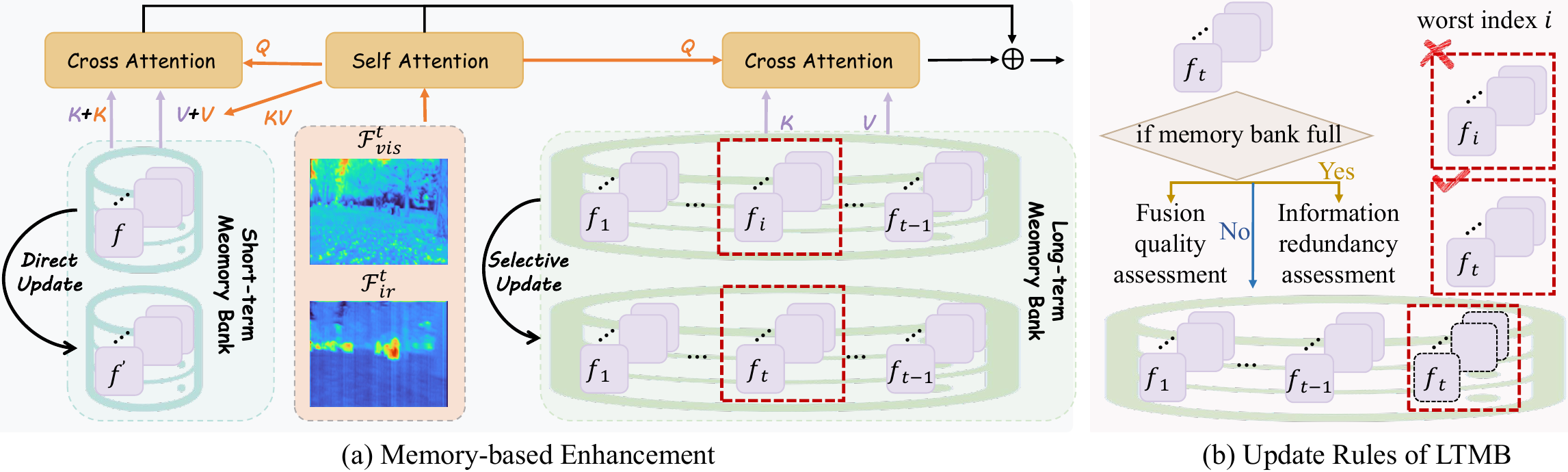}
   \caption{Implementation of memory banks.}
   \label{fig:memory_enhance}
\end{figure}

\subsection{Memory Banks}
\label{Sec:Memory}
For each frame $t$, after obtaining the preliminary fused feature $\mathcal{F}_t$ by fusion layers, we retrieve relevant information from a Short-Term Memory Bank (STMB) and a Long-Term Memory Bank (LTMB) to implement cross-temporal enhancement, as illustrated in Figure~\ref{fig:memory_enhance} (a).

\noindent{\textbf{Short-term Memory Bank (STMB)}} The STMB stores enhanced features from recent frames to provide short-term temporal context for the current fused feature. Since recent frames usually contain locally correlated structures and textures, the short-term memory can provide complementary cues for enhancing dynamic details. Different from directly adding past-frame features to the current representation, our STMB performs current-anchored memory retrieval. Specifically, the current fused feature $\mathcal{F}_t$ provides the query $\mathbf{Q}_c$, as well as the current key and value $(\mathbf{K}_c,\mathbf{V}_c)$, while the STMB provides memory keys and values $(\mathbf{K}_s,\mathbf{V}_s)$. Then the short-term enhanced output is computed via multi-head attention:
\begin{equation}
 \mathbf{tgt}_s=\text{Attention}(\mathbf{Q}_c,\mathbf{K}_c+\mathbf{K}_s,\mathbf{V}_c+\mathbf{V}_s).
\end{equation}

\noindent{\textbf{Long-term Memory Bank (LTMB)}} The LTMB stores representative features from wider-range frames to provide long-term contextual guidance for the current frame. Similar to STMB, the LTMB also adopts current-anchored memory retrieval, where the current fused feature provides the query $\mathbf{Q}_c$, and the LTMB provides memory keys and values $(\mathbf{K}_l,\mathbf{V}_l)$. Different from STMB, the LTMB focuses on retrieving representative long-range contextual cues from temporally distant frames, and is updated with a selective strategy to avoid being dominated by redundant or low-quality historical features. The long-term enhanced output is computed as:
\begin{equation}
 \mathbf{tgt}_l=\text{Attention}(\mathbf{Q}_c,\mathbf{K}_l,\mathbf{V}_l).
\end{equation}
 After both short-term and long-term enhancement, the final enhanced feature is $
 \mathcal{F}^\text{en}_t=\mathcal{F}_t+\mathbf{tgt}_s+\mathbf{tgt}_l$. 
The enhanced feature $\mathcal{F}^\text{en}_t$ is stored in STMB for the next frame enhancement. For LTMB, new features are directly added until the maximum length is reached, and we follow previous work to set the maximum length as 8~\cite{zhou2024rmem}. Otherwise, LTMB follows update rules based on two criteria when receiving a new frame, detailed as follows.

\noindent{\textbf{Update Rules}} 
Although the LTMB effectively stores historical cues for enhancement, its capacity must be restricted to avoid redundant memory accumulation and outdated information. As illustrated in Figure~\ref{fig:memory_enhance} (b), we introduce a dual-criterion pruning mechanism that evaluates each memory entry from two complementary perspectives: \emph{information redundancy}, which measures whether a memory entry can be well reconstructed by the remaining entries, and \emph{fusion quality}, which reflects its ability to perform enhancement and fusion. This design encourages the LTMB to retain diverse and reliable memories, while removing entries that are either redundant or associated with poor fusion quality.

Specifically, when the number of long-term memory entries exceeds the predefined capacity, we perform memory pruning for each batch independently. Given $N$ long-term memory entries $\{\mathbf{M}_i\}_{i=1}^{N}$, we first reshape and normalize them along the feature dimension to obtain $\mathbf{M}\in\mathbb{R}^{N\times d}$. For each memory entry $\mathbf{M}_i$, we use the remaining entries $\mathbf{M}_{-i}$ to linearly reconstruct it by solving a least-squares problem:
\begin{equation}
    \mathbf{w}_i^{*} = \arg\min_{\mathbf{w}_i} 
    \left\| \mathbf{M}_{-i}^{\top}\mathbf{w}_i - \mathbf{M}_i \right\|_2^2 ,
\end{equation}
where $\mathbf{M}_{-i}$ denotes all memory entries except $\mathbf{M}_i$. The reconstructed memory is then obtained as $\hat{\mathbf{M}}_i = \mathbf{M}_{-i}^{\top}\mathbf{w}_i^{*}.$ The reconstruction error is calculated by $e_i = \left\| \hat{\mathbf{M}}_i - \mathbf{M}_i \right\|_2^2.$ A smaller reconstruction error indicates that $\mathbf{M}_i$ can be well represented by other memory entries and is therefore more redundant. Accordingly, we define the reconstruction-based redundancy score as:
\begin{equation}
r_i = 1 - \frac{e_i}{\sum_{j=1}^{N} e_j + \epsilon}.
\end{equation}
In addition to redundancy, we also consider the enhancement and fusion quality. Given the image-level loss $\mathcal{L}_i$ corresponding to the frame stored in the LTMB, we define the quality score as:
\begin{equation}
q_i = \frac{\mathcal{L}_i}{\sum_{j=1}^{N}\mathcal{L}_j + \epsilon},
\end{equation}
where a larger $q_i$ indicates poorer fusion quality and thus a higher priority for removal. Finally, the overall removal score is computed as 
\begin{equation}
  s_i = r_i + q_i.
\end{equation}
The memory entry with the maximum score $i^{*} = \arg\max_i s_i$ is removed, and the LTMB is updated by discarding $\mathbf{M}_{i^{*}}$ while keeping the remaining memory entries. In this way, the LTMB preserves memory entries that are both unique and beneficial to fusion quality, thereby maintaining a compact yet informative long-term representation.

\subsection{Temporal Smoothing Module (TSM)}
To prevent obvious background variations between consecutive frames caused by the integration of differing memory bank features, we introduce a soft adaptive Temporal Smoothing Module (TSM) to implement coarse temporal consistency, which adaptively balances the reliance on current and previous frame information across motion and static regions. Specifically, TSM first estimates a soft motion mask by analyzing inter-frame variations in the original video sequence. Since the infrared modality is often contaminated by noise, which may be erroneously interpreted as motion, the motion estimation is instead performed using the visible video features. Given two consecutive visible-frame features, the difference map is computed as: $D_t = \|\mathcal{F}^{\text{vis}}_t - \mathcal{F}^{\text{vis}}_{t-1}\|_1$, where $\|\cdot\|_1$ denotes the $l_1$ norm. After normalization, a soft motion probability map \(M_t\) is obtained via a sigmoid activation: $M_t = \sigma(norm(D_t))$. To further suppress boundary artifacts and enhance spatial smoothness, the motion map is refined through mean-kernel blurring: $\tilde{M}_t = K * M_t$, where $K$ denotes a mean filter.

Using the resulting soft mask $\tilde{M}_t$, the adaptive smoothing coefficient is constructed as:
\[
\alpha_t = \alpha_{\text{motion}} \cdot \tilde{M}_t + \alpha_{\text{static}} \cdot (1 - \tilde{M}_t),
\]
where \(\alpha_{\text{motion}}\) and \(\alpha_{\text{static}}\) control the degree of temporal filtering for motion and static regions, respectively. Finally, the feature output of the current frame is given by:
\[
\tilde{\mathcal{F}}_t = \alpha_t \odot \mathcal{F}_t + (1 - \alpha_t) \odot \mathcal{F}^{\text{prev}},
\]
where $\tilde{\mathcal{F}}_t$ will serve as the previous frame feature \(\mathcal{F}^{\text{prev}}\) in the computation of the next frame. 

The whole first stage is constrained by temporal-level consistency loss $\mathcal{L}_\text{t}$, and image-level quality loss $\mathcal{L}_\text{grad}, \mathcal{L}_\text{color}$ and $\mathcal{L}_\text{content}$ following~\cite{gong2025temcoco}, concluded as:
\begin{equation}
    \mathcal{L}=\mathcal{L}_\text{t}+\lambda_1\mathcal{L}_\text{grad}+\lambda_2\mathcal{L}_\text{color}+\lambda_3\mathcal{L}_\text{content}.
\end{equation}

\subsection{Temporal Refinement Module (TRM)}
% Although the previous TSM coarsely enhances temporal consistency, the overall temporal consistency of the network is still largely limited by the 2D convolution-based encoder-decoder and other internal structures. Therefore, we design the second-stage Temporal Refinement Module (TSM) to finely improve temporal consistency.

% We assume that the model has already learned high-quality fused feature representations in the first stage. With this powerful fusion prior, the TSM consists of only a few stacked 3D separable convolutions. To further exploit the features, we feed both the first-stage outputs $V_f$ and the original videos $V_{\text{vis}}$ and $V_{\text{ir}}$ into this module and adopt a residual learning strategy to inherit the performance of the first stage. This stage can be mathematically written as:
% \begin{equation}
%     \tilde{V}_f = TRM(V_f, V_{vis}, V_{ir})+V_f,
% \end{equation}
% where $TRM(\cdot)$ denotes the 3D convolutional-based TRM.

% In this stage, we only employ a simple gradient loss $\mathcal{{L}_{\text{grad}}}$ to enhance details and a temporal loss $\mathcal{{L}_\text{t}}$ to reinforce temporal consistency.
Although the previous TSM coarsely enhances temporal consistency, the overall temporal coherence of the network remains largely constrained by the 2D convolution-based encoder-decoder and other spatially oriented structures, which are inherently limited in modeling temporal dynamics. Therefore, we design the second-stage Temporal Refinement Module (TRM) to further refine temporal consistency in a fine-grained and efficient manner.

We assume that the model has already learned high-quality fused feature representations in the first stage. With this strong fusion prior, the TRM is deliberately kept lightweight, consisting of only several stacked 3D separable convolutions to capture subtle temporal variations without excessive computational overhead. To further uncover the informative details overlooked in the first stage, both the fused outputs $V_f$ and the original visible and infrared sequences ($V_{\text{vis}}$, $V_{\text{ir}}$) are fed into this module. A residual learning strategy is employed to inherit and refine the performance of the first stage, leading to a stable and progressive improvement in temporal coherence.

This process can be mathematically formulated as:
\begin{equation}
\tilde{V}_f = TRM(V_f, V_{vis}, V_{ir}) + V_f,
\end{equation}
where $TRM(\cdot)$ denotes the 3D convolution-based refinement network. In this stage, we adopt an optimization objective composed of a gradient loss $\mathcal{L}_{\text{grad}}$ to preserve fine spatial textures and a temporal loss $\mathcal{L}_{\text{t}}$ to enforce inter-frame consistency. This supervision enables the TRM to recover spatio-temporal dynamics overlooked by the first-stage fusion process.

\section{Experiments}
In this section, we first introduce the experimental setup, including the comparison methods, evaluation metrics, hyperparameters, and other implementation details. Subsequently, we present the quantitative and qualitative results on various benchmark datasets. Finally, we perform ablation studies to demonstrate the effectiveness of each designed module in our method.

\subsection{Implementation Details}
\noindent \textbf{Datasets and Metrics} We conduct experiments on two infrared and visible video datasets, namely M3SVD~\cite{tang2025videofusion} and HDO~\cite{xie2024rcvs}, both of which contain pre-registered video pairs.
The M3SVD dataset consists of 220 pairs of video sequences with frame counts ranging from 100 to 1000, while the HDO dataset includes 26 pairs of sequences, each containing 150 to 200 frames.
Our method is trained only on the M3SVD dataset and tested on both datasets.
To comprehensively evaluate the model's performance, we employ MI, CC, SCD and VIF to assess the fusion quality, while flowD and feaCD~\cite{gong2025temcoco} are used to measure the temporal consistency of the fused videos.

\noindent \textbf{Comparison Methods} Since video fusion methods are still in their early stage of development, we select five state-of-the-art image fusion algorithms and three video fusion methods for comparison. The image fusion algorithms include LRRNet~\cite{li2023lrrnet}, Diff-IF~\cite{yi2024diff}, MMIF-EMMA~\cite{zhao2024equivariant}, Text-DiFuse~\cite{zhang2024text}, and TC-MoA~\cite{zhu2024task}, while the video fusion methods include TemCoCo~\cite{gong2025temcoco}, VideoFusion~\cite{tang2025videofusion} and UniVF~\cite{zhao2025unified}. To ensure a fair comparison, for those fusion models lacking integrated de-degradation modules, we apply DSTNet~\cite{rao2024rethinking} to deblur the visible videos and MDIVDNet~\cite{cai2024exploring} to denoise the infrared videos before fusion. Both preprocessing works have been fully retrained on our datasets. All comparative video fusion methods are evaluated using the standard number of frames specified in their official implementations.

\noindent \textbf{Experimental Settings} We empirically set the hyperparameters as $\alpha_{motion}=0.7, \alpha_{static}=0.3, \lambda_1=5, \lambda_2=10$ and $\lambda_3=5$.
During training, the batch size is set to 2, and each batch contains 30 frames. During testing, 200 frames are used for each video sequence. All experiments are conducted on the PyTorch platform using a 2.0 GHz Intel Xeon Gold 6330 CPU and an NVIDIA A800 80 GB PCIe GPU.

%%%%%%%%%%%%%%%%%%%%%%%%%%%%%%%%%%%%%%%%%%%%%%%%%%%%%%%%%%%%

\subsection{Video Fusion}
In this part, we present the experimental results on M3SVD and HDO datasets, including both qualitative and quantitative results, to comprehensively demonstrate the effectiveness of MemoryFusion.

\begin{figure}[t]
  \centering
   \includegraphics[width=1\linewidth]{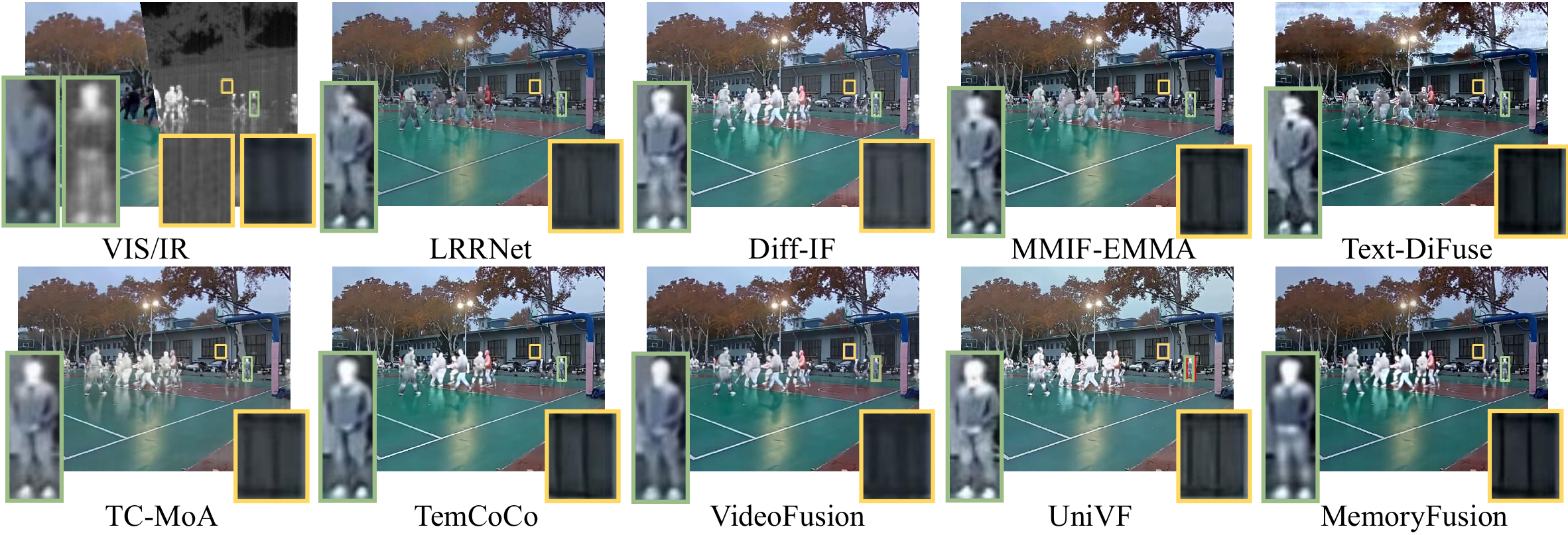}
   \caption{Qualitative results on M3SVD dataset under degradation.}
   \label{fig:results_M3SVD}
\end{figure}

\begin{table*}[t]
\centering
\caption{Quantitative results on the M3SVD and HDO datasets. \textbf{Bold} indicates the best result.}
\small
\setlength{\tabcolsep}{1.5pt}
\begin{tabular}{c|cccccc|cccccc}
\toprule
\multirow{2}{*}{Method} 
& \multicolumn{6}{c|}{M3SVD} 
& \multicolumn{6}{c}{HDO} \\
\cmidrule(lr){2-7} \cmidrule(lr){8-13}
& MI$\uparrow$ & CC$\uparrow$ & SCD$\uparrow$ & VIF$\uparrow$ & flowD$\downarrow$ & feaCD$\downarrow$
& MI$\uparrow$ & CC$\uparrow$ & SCD$\uparrow$ & VIF$\uparrow$ & flowD$\downarrow$ & feaCD$\downarrow$ \\
\midrule
LRRNet 
& 1.7889 & 0.4584 & 1.4884 & 0.3906 & 5.5065 & 1.9309
& 1.0616 & 0.6725 & 0.7250 & 0.2416 & 7.4598 & 1.9523 \\
Diff-IF 
& 2.0692 & 0.4912 & 1.6068 & 0.4847 & 4.0890 & 1.8210
& 1.1387 & 0.6851 & 1.1483 & 0.3039 & 8.0450 & 1.9587 \\
MMIF-EMMA 
& 2.1175 & 0.4758 & 1.6443 & 0.4758 & 4.6592 & 1.8412
& 1.1668 & 0.6945 & 1.2582 & 0.3071 & 7.9598 & 1.9565 \\
Text-DiFuse 
& 1.6253 & 0.4576 & 1.4989 & 0.3514 & 8.6578 & 1.9744
& 1.1682 & 0.7024 & 1.4096 & 0.2940 & 9.7262 & 1.9732 \\
TC-MoA 
& 2.0178 & 0.4660 & 1.5276 & 0.4054 & 4.1954 & 1.8118
& 1.0942 & 0.7023 & 1.1639 & 0.3181 & 7.8125 & 1.9570 \\
TemCoCo 
& 2.1717 & 0.4800 & 1.5961 & 0.4770 & 3.1882 & 1.8058
& 1.3212 & 0.7192 & 1.3689 & 0.3823 & 5.1147 & 1.7336 \\
VideoFusion 
& 2.0353 & 0.4612 & 1.5299 & 0.4126 & 3.4161 & 1.8154
& 1.1428 & 0.7106 & 1.3152 & 0.3238 & 6.0564 & 1.8643 \\
UniVF 
& 2.1316 & 0.4848 & 1.6122 & 0.4677 & 4.0271 & 1.8138
& 1.1854 & 0.7205 & 1.3266 & 0.3427 & 6.7987 & 1.8680 \\
\midrule\\[-1em]
Ours-stage1 
& 2.1631 & 0.4872 & 1.5975 & 0.4877 & 3.6135 & 1.8088
& 1.3261 & 0.7319 & 1.4002 & 0.3733 & 6.6011 & 1.7747 \\
MemoryFusion 
& \textbf{2.2350} & \textbf{0.4952} & \textbf{1.6152} & \textbf{0.4926} & \textbf{2.6759} & \textbf{1.7640}
& \textbf{1.3264} & \textbf{0.7347} & \textbf{1.4450} & \textbf{0.3947} & \textbf{4.9188} & \textbf{1.7293} \\
\bottomrule
\end{tabular}
\label{tab:quantitative_results}
\end{table*}

\noindent \textbf{Results on M3SVD}
Figure~\ref{fig:results_M3SVD} compares different methods on the M3SVD dataset. As shown, the proposed MemoryFusion exhibits clear advantages in both detail restoration and contrast preservation. Specifically, frame-based fusion methods fail to preserve complete and realistic human structures or suffer from noticeable detail blurring. Among video-based methods, TemCoCo and VideoFusion remain limited by low contrast or misalignment artifacts, while UniVF, despite sharper structures, is still inferior. In the yellow box at the lower right, our method reconstructs window structures with the highest contrast among all competitors.
Table~\ref{tab:quantitative_results} reports the quantitative results on M3SVD. Our method achieves the best performance on all image quality metrics (MI, CC, SCD, VIF) as well as temporal consistency metrics (flowD, feaCD), demonstrating superior spatial fidelity and temporal coherence. While TemCoCo and VideoFusion show relatively strong temporal consistency due to video-sequence modeling, UniVF exhibits weaker temporal coherence without such mechanisms. Notably, our first-stage model already surpasses image-based methods due to temporal consistency loss and TSM, though frame-wise processing limits its temporal performance. With the lightweight second stage, temporal coherence is effectively restored and image quality is further improved.

\begin{figure}[t]
  \vspace{-0.2in}
  \centering
   \includegraphics[width=1\linewidth]{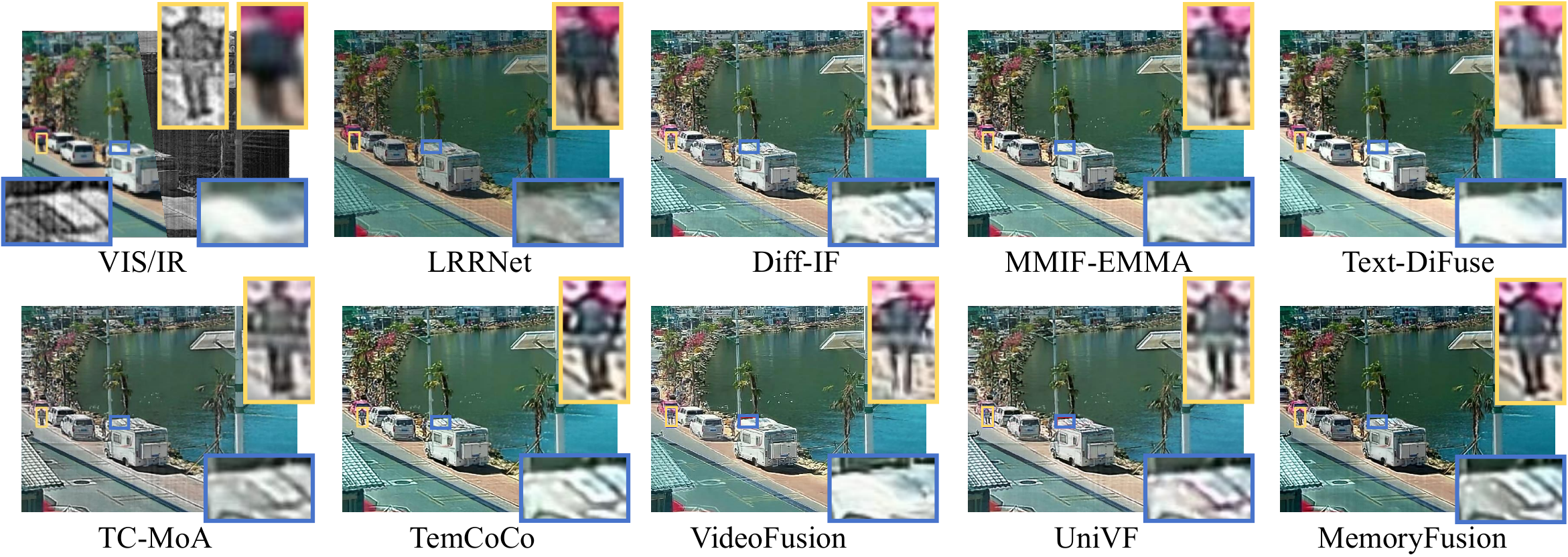}
   \caption{Qualitative results on HDO dataset under degradation.}
   \label{fig:results_HDO}
\end{figure}

\noindent \textbf{Results on HDO}
Figure~\ref{fig:results_HDO} presents the qualitative results on the HDO dataset. In the enlarged yellow box at the upper right, similar to the previous results, all compared methods except TemCoCo and our proposed MemoryFusion fail to preserve the complete human structure, where TC-MoA and UniVF exhibit noticeable color distortion, and Diff-IF, MMIF-EMMA and VideoFusion produce blurred or fragmented contours.
In contrast, both TemCoCo and our approach maintain more coherent human outlines, yet TemCoCo still suffers from residual blur and local detail loss.
Beyond the structural regions, the enlarged blue box at the lower right highlights differences in fine texture reconstruction. Our MemoryFusion yields sharper edges and more faithful contrast, effectively restoring spatial details that are heavily degraded in other methods.
The quantitative results summarized in Table~\ref{tab:quantitative_results} further confirm these observations, showing that MemoryFusion consistently achieves superior scores across all metrics, validating its robustness and effectiveness under severe degradation conditions.

% Figure~\ref{fig:results_HDO} presents the qualitative results on the HDO dataset. In the enlarged yellow box at the upper right, similar to the previous results, all compared methods except TemCoCo and our proposed MemoryFusion fail to preserve the complete human structure, where TC-MoA and UniVF exhibit noticeable color distortion, and Diff-IF, MMIF-EMMA, and VideoFusion produce blurred or fragmented contours. This indicates that severe degradation not only suppresses local appearance cues but also makes it difficult for frame-wise or locally modeled methods to maintain coherent object structures across time. In contrast, both TemCoCo and our approach maintain more coherent human outlines, yet TemCoCo still suffers from residual blur and local detail loss.
% Beyond the structural regions, the enlarged blue box at the lower right highlights differences in fine texture reconstruction. Our MemoryFusion yields sharper edges and more faithful contrast, effectively restoring spatial details that are heavily degraded in other methods. 
% The quantitative results summarized in Table~\ref{tab:quantitative_results} further confirm these observations, showing that MemoryFusion consistently achieves superior scores across all metrics, validating its robustness and effectiveness under severe degradation conditions.

\begin{wrapfigure}{r}{0.55\textwidth}
  \centering
  \vspace{-0.5em}
  \includegraphics[width=\linewidth]{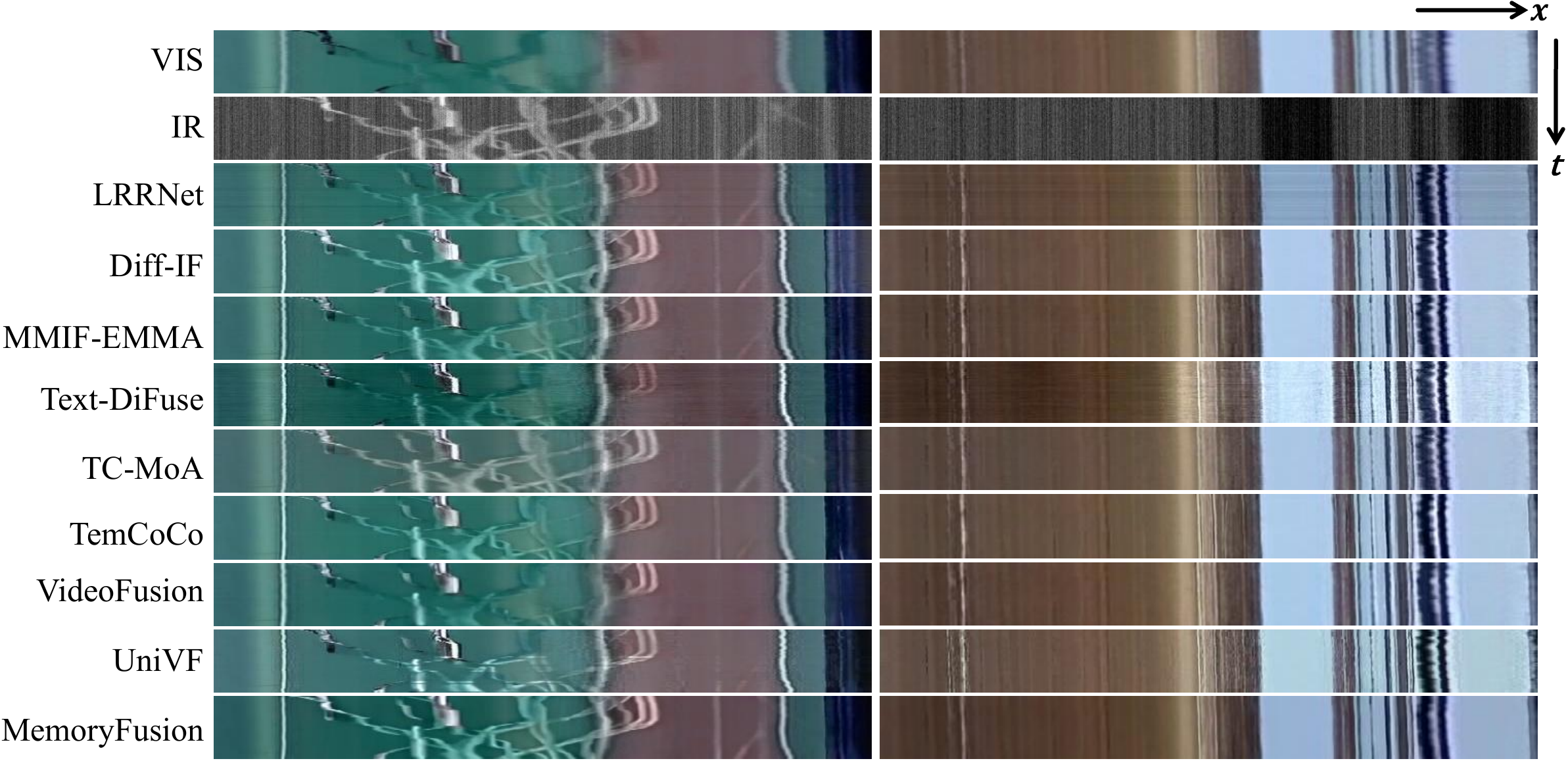}
  \caption{Temporal consistency comparison.}
  \label{fig:results_temporal}
  \vspace{-0.8em}
\end{wrapfigure}

\noindent \textbf{Temporal Consistency Comparison}
To evaluate temporal consistency, we concatenate the same pixel row from 60 consecutive frames for visualization, as shown in Figure~\ref{fig:results_temporal}. Existing methods suffer from temporal artifacts. For instance, LRRNet and Text-DiFuse show ghosting and tearing effects, while Diff-IF and MMIF-EMMA exhibit structural discontinuities. Although TemCoCo, VideoFusion, and UniVF reduce some distortions via frame alignment, they still struggle with smooth temporal transitions. In contrast, MemoryFusion produces temporally coherent and visually stable results with well-preserved structures, demonstrating its effectiveness in modeling temporal dependencies and maintaining spatial-temporal stability.

% \begin{figure}[t]
%   \centering
%    \includegraphics[width=1\linewidth]{figs/temporal_compare.pdf}
%    \caption{Temporal consistency comparison.}
%    \label{fig:results_temporal}
% \end{figure}

\subsection{Ablation Studies}
\begin{figure}[t]
  \vspace{-0.2in}
  \centering
   \includegraphics[width=1\linewidth]{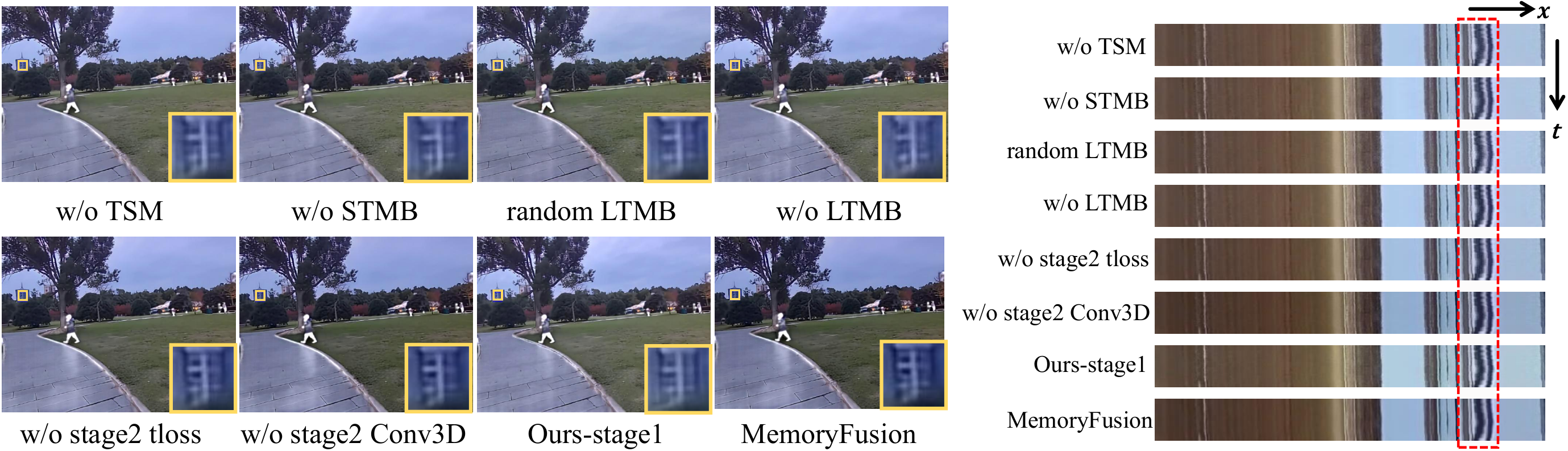}
   \caption{Qualitative results of ablation experiments.}
   \label{fig:results_ablation}
\end{figure}

In this section, we conduct comprehensive ablation studies to validate the effectiveness of each designed component in MemoryFusion. Specifically, we provide qualitative evaluations from both image quality and temporal consistency perspectives, as shown in Figure~\ref{fig:results_ablation}, and report the corresponding quantitative results in Table~\ref{tab:ablation}. It is worth noting that, except for the two variants related to the second refinement stage (\emph{i.e.}, w/o stage2 tloss and w/o stage2 Conv3D), all other ablation settings are evaluated only within the first stage. Therefore, these variants are compared with the stage-1 baseline rather than the complete MemoryFusion model, so as to more accurately reflect the contribution of each stage-1 component.

\noindent \textbf{w/o TSM} This setting removes the Temporal Smoothing Module (TSM). The TSM leverages the previous frame to smooth the current frame, reducing obvious inconsistency between consecutive frames. As shown in Figure~\ref{fig:results_ablation} and Table~\ref{tab:ablation}, omitting this module has little effect on image quality but results in temporal flickering, confirming its role in enhancing temporal smoothness. This indicates that TSM mainly contributes to inter-frame stability.

\noindent \textbf{w/o STMB} This experiment disables the Short-Term Memory Bank (STMB). Without it, the model cannot effectively reuse short-term contextual cues from recent frames to guide current-frame fusion. As shown in the figure, the result shows blurred textures and weakened local contrast. These results confirm that STMB plays a vital role in reinforcing details during enhancement and fusion. This suggests that recent temporal information is beneficial for preserving local structures and fine textures.

\noindent \textbf{Random LTMB} In this configuration, the Long-Term Memory Bank (LTMB) is updated randomly rather than by the proposed dual-criterion rule. Such stochastic replacement disrupts the accumulation of informative long-term cues, leading to unstable global guidance. The results highlight the importance of an effective update mechanism for maintaining meaningful long-term memory. This further shows that the benefit of LTMB comes not only from memory storage, but also from selective memory updating.

\noindent \textbf{w/o LTMB} This variant removes the LTMB to examine the contribution of global memory aggregation. Without access to long-term frame information, the model is limited to short-range temporal cues from the STMB and fails to capture global contextual relations and scene structures. As a result, the fused results exhibit degraded structural integrity and weakened detail representation, especially in complex or spatially diverse regions, demonstrating the effectiveness of LTMB. This verifies that long-term memory provides complementary global cues beyond short-term temporal memory.

\begin{wraptable}{r}{0.65\textwidth}
\centering
\caption{Quantitative results of ablation experiments.}
\label{tab:ablation}
\small
\setlength{\tabcolsep}{2pt}
\renewcommand{\arraystretch}{0.9}
\begin{tabular}{c|cccccc}
\toprule
M3SVD & MI$\uparrow$ & CC$\uparrow$ & SCD$\uparrow$ & VIF$\uparrow$ & flowD$\downarrow$ & feaCD$\downarrow$ \\
\midrule
w/o TSM & 2.0173 & 0.4821 & 1.5734 & \textbf{0.4943} & 4.2172 & 1.8038 \\
w/o STMB & 2.0239 & 0.4767 & 1.5538 & 0.4817 & 3.6586 & 1.8061 \\
random LTMB & 2.0199 & 0.4755 & 1.5580 & 0.4825 & 3.7010 & 1.8036 \\
w/o LTMB & 2.0031 & 0.4646 & 1.5509 & 0.4770 & 3.7768 & 1.8078 \\
w/o stage2 tloss & 2.1987 & 0.4884 & 1.5961 & 0.4932 & 3.2065 & 1.7985 \\
w/o stage2 Conv3D & 2.1007 & 0.4942 & 1.5994 & 0.4938 & 3.5290 & 1.7956 \\
\midrule
Ours-stage1 & 2.1631 & 0.4872 & 1.5975 & 0.4877 & 3.6135 & 1.8088 \\
MemoryFusion & \textbf{2.2350} & \textbf{0.4952} & \textbf{1.6152} & 0.4926 & \textbf{2.6759} & \textbf{1.7640} \\
\bottomrule
\end{tabular}
\end{wraptable}
%These findings demonstrate that the LTMB is crucial for supplying global feature guidance and structure reinforcement, which effectively enhances the representational richness and overall fusion quality of the model.

% \noindent \textbf{w/o stage2 tloss \& w/o stage2 Conv3D} These two variants evaluate the contribution of the temporal loss and the 3D convolution in the second stage. Removing the temporal loss weakens temporal smoothness and slightly reduces fusion quality, while removing the 3D convolution causes a more pronounced decline in both temporal coherence and perceptual fidelity. As shown in Table~\ref{tab:ablation}, the 3D convolution plays a more decisive role in maintaining temporal stability, demonstrating that the joint spatio-temporal refinement in stage 2 is crucial for stabilizing video fusion results.

\noindent \textbf{w/o stage2 tloss \& w/o stage2 Conv3D}
These two variants evaluate the temporal loss and 3D convolution in the second refinement stage. Removing the temporal loss weakens the explicit supervision on inter-frame consistency, leading to less stable transitions. For the w/o stage2 Conv3D variant, we replace the 3D convolution with 2D convolution, which preserves spatial refinement but removes explicit temporal modeling across adjacent frames. As shown in Table~\ref{tab:ablation}, 3D convolution plays a more decisive role in maintaining temporal stability, demonstrating that the joint spatio-temporal refinement in the second stage is crucial for stabilizing video fusion results.

\subsection{Plug-and-Play Spatiotemporal Refinement with TRM}
\label{sec:trm_generalization}

To further investigate the generalization and effectiveness of the proposed TRM, we apply it to two existing fusion methods, MMIF-EMMA and UniVF. Specifically, the original fusion models are kept fixed, and their fused outputs are directly used as fusion priors for TRM. This allows TRM to perform spatiotemporal refinement without modifying the original fusion networks, demonstrating its flexibility in enhancing existing fusion methods.

As shown in Table~\ref{tab:trm_generalization}, introducing TRM significantly improves the temporal consistency of both methods, achieving relative reductions in flowD of 36.6\% and 30.4\% for MMIF-EMMA and UniVF, respectively. In addition to these substantial improvements in temporal consistency, TRM also enhances fusion quality in terms of CC and SCD for both methods, while maintaining comparable performance on other fusion metrics. These results demonstrate that TRM can effectively enhance temporal consistency while preserving or improving spatial fusion quality, even when applied to fusion models without memory mechanisms.

\begin{wraptable}{r}{0.6\textwidth}
\centering
\caption{Quantitative comparison of applying TRM to different fusion methods. The best results are highlighted in bold.}
\label{tab:trm_generalization}
\scriptsize
\setlength{\tabcolsep}{3pt}
\renewcommand{\arraystretch}{0.9}
\begin{tabular}{l|cccccc}
\toprule
Method & MI$\uparrow$ & CC$\uparrow$ & SCD$\uparrow$ & VIF$\uparrow$ & flowD$\downarrow$ & feaCD$\downarrow$ \\
\midrule
MMIF-EMMA
& 2.1175 & 0.4758 & 1.6443 & 0.4758 & 4.6592 & 1.8412 \\
MMIF-EMMA+TRM
& 2.0641 & 0.4932 & \textbf{1.6624} & 0.4762 & 2.9549 & 1.8125 \\
\midrule
UniVF
& 2.1316 & 0.4848 & 1.6122 & 0.4677 & 4.0271 & 1.8138 \\
UniVF+TRM
& 2.0337 & 0.4921 & 1.6325 & 0.4673 & 2.8025 & 1.8122 \\
\midrule
Ours-stage1
& 2.1631 & 0.4872 & 1.5975 & 0.4877 & 3.6135 & 1.8088 \\
\textbf{MemoryFusion}
& \textbf{2.2350} & \textbf{0.4952} & 1.6152 & \textbf{0.4926} & \textbf{2.6759} & \textbf{1.7640} \\
\bottomrule
\end{tabular}
\end{wraptable}

Furthermore, the complete MemoryFusion achieves the best temporal consistency and superior overall fusion performance, suggesting that the memory-enhanced Stage-1 fusion prior provides a favorable foundation for subsequent refinement. More importantly, TRM achieves substantial improvements when directly applied to other fusion methods without requiring additional training or fine-tuning, highlighting its plug-and-play capability. These results demonstrate the general applicability of TRM as a flexible spatiotemporal refinement module that can effectively enhance fusion quality and, more importantly, temporal consistency, with the potential for further improvements through finetuning.

\subsection{Computation Efficiency Comparison}
We compare the computational efficiency in terms of FLOPs, parameter count, and runtime in Table~\ref{tab:efficiency}. Although MemoryFusion introduces an additional second stage, the overall computational cost remains moderate. Specifically, the first-stage model achieves competitive efficiency compared with existing video-based fusion methods, while the lightweight second stage incurs only negligible overhead. For preprocessing-based methods, the reported values only account for the computational cost of the fusion networks, excluding additional pretrained deblurring or denoising modules. Therefore, the comparison reflects the intrinsic efficiency of the fusion networks rather than the computational cost of the complete preprocessing-and-fusion pipelines. Overall, our full model strikes a favorable balance between fusion quality and computational efficiency, outperforming heavy diffusion-based or large-scale temporal models while maintaining comparable inference speed.

\begin{table}[htbp]
    \centering
    \caption{Comparison of computational efficiency.}
    \label{tab:efficiency}
    \renewcommand{\arraystretch}{1}
    \setlength{\tabcolsep}{1.2pt}   
    % \scriptsize
    \footnotesize
    
    \begin{tabular}{lcccccccc|cc}
    \toprule
% &  &  &  &  &  &  &  &   & \multicolumn{2}{c}{\textbf{Ours}} \\
% \cmidrule(lr){10-11}
 & \text{LRRNet} & \text{Diff-IF} & \text{MMIF-EMMA} & \text{Text-DiFuse} & \text{TC-MoA} & \text{TemCoCo} & \text{VideoFusion} & \text{UniVF} &  Stage1 &  Stage2 \\

\midrule
    FLOPs (G)  & 14.2  & 86.9   & 41.5  & 1268 & 4833  & 505.6 & 301.1 & 740.7 & {213.6} & {1.65} \\
    Param (M) & 0.047    & 22.36  & 1.52  & 119.4  & 340   & 19.18 & 6.74  & 9.15  & {17.2}  & {0.033} \\
    Time (s)              & 0.162  & 4.23   & 0.056  & 24.2  & 0.185  & 0.109  & 0.105  & 0.245  & {0.099}  & {0.001} \\
    \bottomrule
    \end{tabular}
\end{table}

\section{Conclusion}

This paper proposes MemoryFusion, a two-stage cross-temporal multimodal video fusion framework designed to achieve high-fidelity fusion and superior temporal consistency. In the first stage, dual memory banks dynamically update and effectively retrieve informative cues from historical frames to enhance current-frame fusion. Additionally, a TSM is incorporated to adaptively mitigate coarse frame variations and establish a stable temporal foundation. In the second stage, guided by the powerful fusion prior from the first stage, a lightweight TRM further enforces fine-grained temporal coherence and enhances spatial fidelity through residual refinement. Extensive experiments on multiple public datasets demonstrate that MemoryFusion not only achieves state-of-the-art performance in both fusion quality and temporal consistency but also exhibits strong generalization across diverse scenes.

\begin{ack}
This work was supported by the National Natural Science Foundation of China under Grant no. 62676289, and the China Postdoctoral Science Foundation under no. 2026T190426. The authors declare no competing interests.
\end{ack}

{\small
 \bibliographystyle{plainnat}
 \bibliography{egbib}
}

\appendix
\clearpage
\section{Supplemental Material}

\subsection{Experimental Details}
MemoryFusion directly takes the raw infrared-visible video sequences as input without using external de-degradation preprocessing. For baseline methods that do not contain degradation-removal components, we additionally apply DSTNet/MDIVDNet as preprocessing only for robustness evaluation under degraded inputs. Unless otherwise stated, the FLOPs, parameters, and inference time reported for these baseline fusion networks do not include the additional computational cost of DSTNet/MDIVDNet. Therefore, the efficiency comparison is favorable to the baselines, while MemoryFusion remains competitive without relying on external preprocessing.

\subsection{Extension to Medical Scenes}
\begin{figure}[h]
  \renewcommand\thefigure{s1}
  \centering
   \includegraphics[width=1\linewidth]{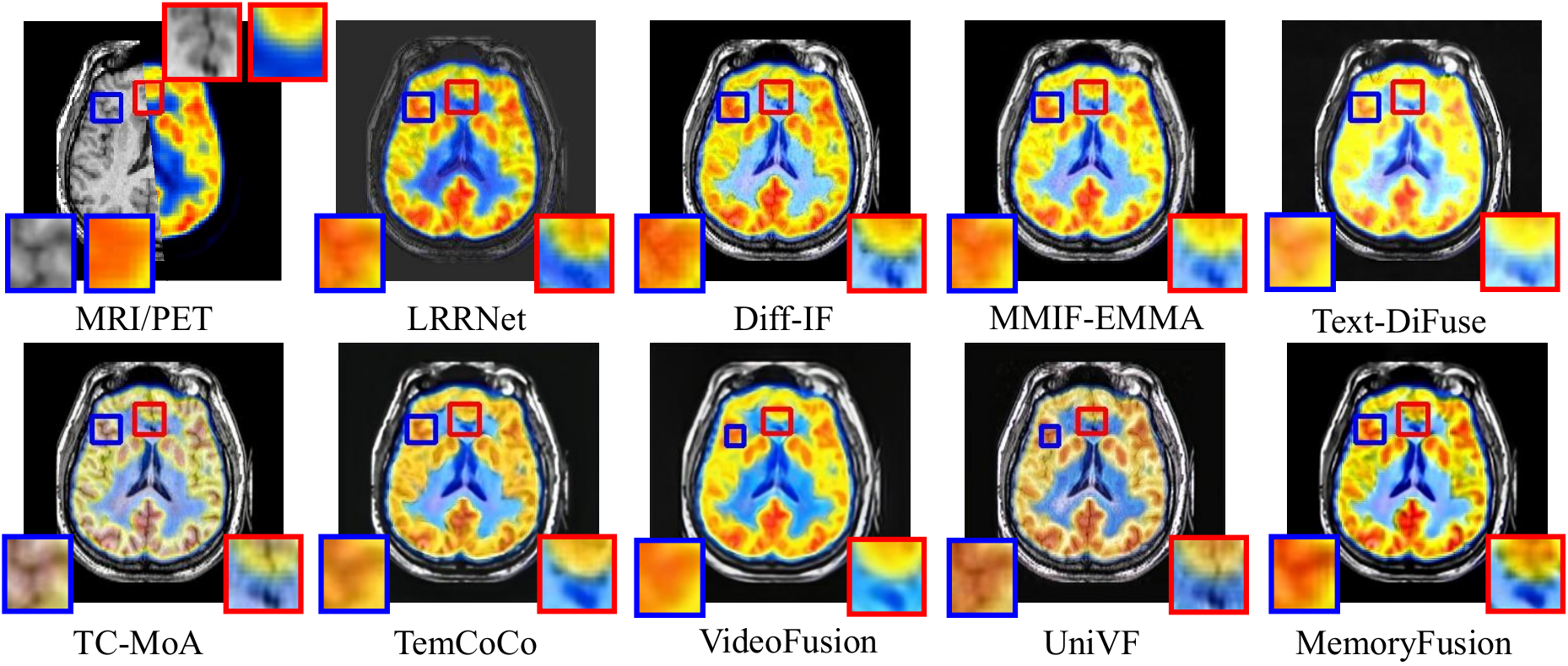}
   \caption{Results of medical sequence fusion on Harvard dataset.}
   \label{fig:results_medical}
\end{figure}

We further extend the proposed model to medical sequence fusion to evaluate its generalization capability to medical scenes. Specifically, experiments are conducted on the Harvard medical imaging dataset\footnote{\url{http://www.med.harvard.edu/AANLIB/}}, where the MRI and PET sequences consist of a series of consecutive slice images acquired from cross-sectional scans of the same organ. Owing to the strong anatomical correlation between adjacent slices, the spatial continuity among them is analogous to the temporal coherence between consecutive frames in a video. Hence, we treat each MRI or PET slice as a video frame to evaluate the model's performance in medical sequence fusion. Since medical scenes are relatively homogeneous, this experiment primarily focuses on assessing the fusion performance. As shown in Figure~\ref{fig:results_medical}, methods such as LRRNet, Text-DiFuse and VideoFusion tend to blur fine structural boundaries or over-smooth functional details, while Diff-IF and MMIF-EMMA exhibit noticeable inconsistencies between structural and metabolic regions. Despite their competence in detail preservation, TC-MoA and UniVF fail to retain the color information of PET images, resulting in severe spectral distortion. In contrast, our MemoryFusion framework effectively integrates structural information from MRI with metabolic patterns from PET, achieving both high structural fidelity and functional contrast. The highlighted regions further demonstrate that our approach preserves sharp anatomical edges and detailed texture in gray matter areas while maintaining accurate PET intensity distributions, thereby providing more clinically interpretable fused results.

\subsection{Results of Non-degraded Scenes}
\begin{figure}[hb]
  \renewcommand\thefigure{s2}
  \centering
   \includegraphics[width=1\linewidth]{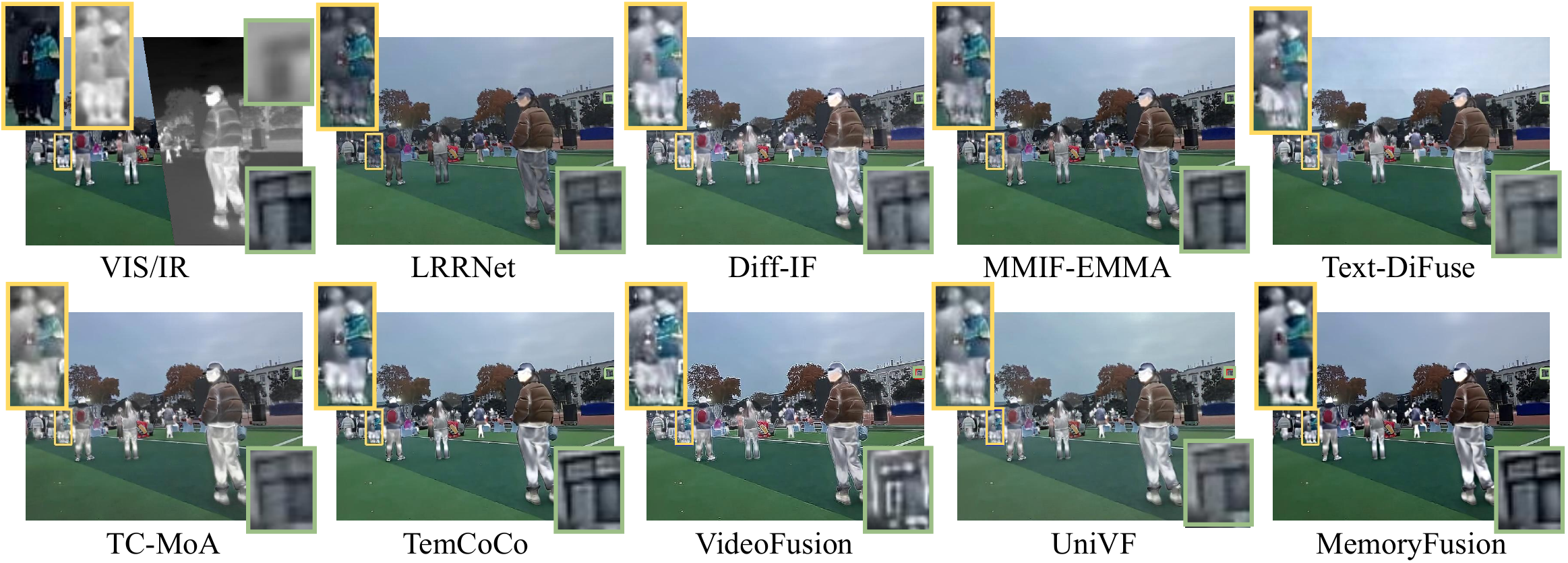}
   \caption{Qualitative results of non-degraded scene.}
   \label{fig:results_M3SVD_enhance}
\end{figure}

Figure~\ref{fig:results_M3SVD_enhance} shows the qualitative results on non-degraded scenes. MemoryFusion enhances visual contrast and salient structural cues. In the enlarged yellow box at the upper left, most methods fail to clearly separate the human from the background, while TemCoCo and VideoFusion introduce boundary artifacts due to over-enhancement. In contrast, MemoryFusion effectively incorporates thermal cues from the infrared modality, producing clearer human boundaries with well-preserved texture continuity. The enlarged green box at the lower right further compares texture reconstruction around the window region. Obviously, our MemoryFusion delivers sharper and more consistent structures with higher contrast. These results demonstrate that the proposed memory-enhanced modeling not only preserves global scene semantics but also improves fine-detail reconstruction, even under non-degraded conditions where other methods tend to over-smooth or distort details.

\subsection{Limitation}
Although MemoryFusion achieves strong fusion quality and temporal consistency, it is not explicitly designed as a degradation-specific restoration framework. Its robustness thus depends partly on the degradation coverage of the training data, and may decrease under severely unseen degradations. Moreover, since TSM uses visible-frame differences only as a lightweight temporal smoothing cue, when motion is mainly reflected in the infrared modality, the first-stage smoothing may provide limited guidance, placing greater pressure on the second-stage Temporal Refinement Module to restore temporal consistency. Nevertheless, these limitations are orthogonal to our core contribution, and degradation-aware modeling or reliability-aware bi-modal motion estimation can be naturally incorporated into MemoryFusion as additional modules in future work.

% \newpage
% \input{checklist.tex}

\end{document}